\documentclass[letterpaper, 10 pt, conference]{ieeeconf}  

\IEEEoverridecommandlockouts                              

\usepackage[utf8]{inputenc}
\usepackage[T1]{fontenc}
\usepackage{algorithm}
\usepackage{algpseudocode}
\usepackage[separate-uncertainty = true]{siunitx}
\usepackage{caption}
\usepackage{graphicx}

 \newenvironment{Figure}
  {\par\medskip\noindent\minipage{\linewidth}}
  {\endminipage\par\medskip}
 
\usepackage{biblatex}
\title{\LARGE \bf
Learning Actions and Control of Focus of Attention with a Log-Polar-like Sensor
}

\author{Robin Göransson$^{1}$, and Volker Krueger$^{1}$
	\thanks{*This work was partially supported by the Wallenberg AI, Autonomous Systems and Software Program (WASP) funded by Knut and Alice Wallenberg Foundation.}
	\thanks{$^{1}$Department of Computer Science, Faculty of Engineering (LTH), Lund University, SE~221~00 Lund, Sweden. E-mail: <firstname>.<lastname>@cs.lth.se.
	}%
}

\begin{document}

\maketitle
\thispagestyle{empty}
\pagestyle{empty}

\begin{abstract}
With the long-term goal of reducing the image processing time on an autonomous mobile robot in mind we explore in this paper the use of log-polar like image data with gaze control. The gaze control is not done on the Cartesian image but on the log-polar like image data. For this we start out from the classic deep reinforcement learning approach for Atari games. We extend an A3C deep RL approach with an LSTM network, and we learn the policy for playing three Atari games and a policy for gaze control. While the Atari games already use low-resolution images of $\mathbf{80\times80}$ pixels,  we are able to further reduce the amount of image pixels by a factor of 5 without losing any gaming performance.
\end{abstract}

\section{Introduction}
One goal of autonomous robots is to let them act autonomously over an extended period of time without human intervention. The autonomous robots would use processed video input to decide which action they should take. 
Small scale autonomous robots such as drones have limited battery power and the necessary on-line processing of the video data has an impact on their runtime. 
One classic idea to reduce the image processing resources is to take inspiration from the human visual system (HVS): the human eyes famously have a non-homogeneous resolution, with a high-resolution fovea in the center and with a linearly decreasing resolution towards the periphery~\cite{curcio90}.  This resolution comes with the challenge of deciding where to focus the attention, i.e., what part of the field of view should be foveated. The HVS employs a process that continuously controls the gaze. Many computer vision systems exploit the idea of controlling the focus of attention by computing salient feature points. This usually comes at a computational cost because the processing is done on the Cartesian image. The gaze control of the HVS, however, is based on the low resolution retinal image, which implies lower computational needs~\cite{curcio90,foa}.

In this paper we explore first steps towards the idea of a) controlling a robot based on a retinal image and b) controlling the gaze based on the retinal image as well, i.e., we want to use only retinal image data at every point in the artificial visual pathway.

In detail, the starting point of our exploration is the classic deep RL approach for playing Atari games~\cite{mnih90nature}. We revised the original approach by using the S.o.t.A. deep RL approach A3C~\cite{a3c} for learning two policies: one for controlling the movements of the robot player and one for controlling the gaze.  The input to the deep RL process are log-polar-like images that are computed directly from the Cartesian Atari images. As the smaller log-polar-like images made state-estimation harder, we extended the A3C approach with an LSTM~\cite{lstm} and Generalized Advantage Estimation (GAE)~\cite{gae}.
We have explored different log-polar-like sensor sizes and resolutions.
We have evaluated the performance on three different Atari games: Pong, Breakout and Beam Rider, with Pong being the simplest and Beam Rider the most challenging one of the three. The experiments for Pong and Breakout show that we can reduce the image data by a factor of 5 while still achieving similar game performance. The learning of the two policies took approx. 5 times longer, but since this is done off-line we do not consider this to be a problem.
As the results for Pong and Breakout were similar, we will report only the Breakout results in detail. In case of Beam Rider we were not able to complete policy training as the learning took considerably longer. However, intermediate results were similar to Pong and Breakout.

In summary, our contributions are:
\begin{itemize}
    \item We explored the use of log-polar like image data and gaze control to reduce image computation time.
    \item We used an A3C deep RL approach, extended it with an LSTM network and GAE, and we used this to train two policies, one for playing the game and one for controlling the gaze.
    \item We only used the log-polar like image data throughout the entire artificial visual pathway.
    \item We evaluated this on three Atari games and the results hint that we can reduce the image data by a factor of approx. 5 without losing performance.
\end{itemize}

\section{Background and Related Work}

The ability to control gaze direction allows to efficiently sample the visual field of view~\cite{elazary08jv,lombardi02springer}. This should reduce computation time. In computer vision, salient features are often used to focus in on relevant image areas~\cite{wang19cvpr,ruan13cvpr,Liu12csae,hou07cvpr,ming11cvpr,orabona07}. 
The idea of log-polar sensors is frequently used for a variety of applications incl. navigation~\cite{traver10RAS,tistarelli93pami}, object recognition ~\cite{billandon20,electronics10222883,sandini90a,amorim18ijcnn,daniilinidis18iclr} and motion computation~\cite{daniilinidis95caip,li17electron,sandini90a}. See \cite{traver10RAS} for a good overview. 
Most publications that use (log-)polar transforms aim to exploit the rotational invariance and the simplified use of scale that comes with that transform. Recent work here includes ~\cite{daniilinidis18iclr,electronics10222883,billandon20,hu21nc,amorim18ijcnn}.

For better readability of our paper we provide below a brief overview of the employed deep RL techniques.

\textbf{A3C}: Combining reinforcement learning, where updates usually are heavily correlated, and deep learning, where the input samples generally are assumed to be independent, can be problematic. The Asynchronous Advantage Actor-Critic (A3C) algorithm \cite{a3c} is a deep reinforcement learning algorithm that decorrelates updates by asynchronously executing multiple agents in parallel. Each agent uses its own set of network parameters and local copy of the environment which means that at every given time-step the agents experience a variety of different states and consecutive updates will no longer be heavily correlated \cite{a3c}. \\

The A3C has two outputs: a softmax output for the policy $\pi_\theta(s)$ and a linear output for the value function $V_\theta(s)$. The value function can be used to calculate the advantage function as $A_\theta(s_t) = (\sum_{i=0}^{k-1}\gamma^i R_{t+1+i} + \gamma^k V_\theta(s_{t+k})) - V_\theta(s_t)$ where $s_t$ is the state for time-step $t$, $R_{t+1}$ is the reward received when taking action $a_t$ in state $s_t$ and $\gamma$ is the discount factor \cite{a3c}. With this advantage function the value loss is defined as $L_{value} = \frac{1}{2}\sum_{i=0}^{k-1}  A_\theta(s_{t+i})^2$ and the policy loss as $L_{policy} = \sum_{i=0}^{k-1}(-\mathrm{log}\pi_\theta(a_{t+i}|s_{t+i})A_\theta(s_{t+i})-\beta H(\pi_\theta(s_{t+i})))$ where $H(\pi_\theta) = -\sum \pi_\theta \log \pi_\theta$ is an entropy term and $\beta$ controls the strength of this entropy term \cite{a3c}. When an agent reaches a terminal state, or has performed $t_{max}$ steps, the local gradient is computed using the total loss. This local gradient is then applied to the global network in an asynchronous fashion and the local network is reinitialized using the now updated global parameters. \\

Exploration vs. exploitation is an important concept for a reinforcement learning algorithm. The softmax policy output of the A3C can be seen as a probability distribution over the actions in each state and the next action can thus be chosen probabilistically \cite{a3c}. Due to the nature of the softmax function every action will always have a non-zero probability of being chosen, making it possible for the algorithm to explore the full state-space. \\

\textbf{A3C-LSTM}: In the Atari environments, and in many robotics environments as well, consecutive states are correlated and more than one state is needed to determine which way something is moving. In order to give the agent access to information from more than the current frame we introduce recurrent elements to the network. More specifically, a Long Short-Term Memory (LSTM) \cite{lstm} module is added to the network after the convolutional layers. The LSTM is fed with the output from the convolutional layers and the output from the LSTM from the previous time-step. \\

\textbf{GAE}: Generalized Advantage Estimation (GAE) is used to make a more robust estimation of the advantage function \cite{gae}. Different TD-errors (1-step, 2-step, ..., $k$-step) are used to form advantage estimators $\hat{A}_t^{(k)}$. The generalized advantage estimator $\hat{A}_t^{GAE}$ is then defined as the exponentially-weighted average of the different estimators $\hat{A}_t^{(k)}$. When forming the average a parameter $0\leq \lambda \leq 1$ is used. This parameter works as a trade-off between bias and variance: for $\lambda=1$ the variance is usually high while a lower $\lambda$ reduces the variance but introduces bias \cite{gae}.
\section{Focus-of-Attention}
When an agent is trained on an Atari environment full pre-processed screens are used as input. In this paper the input is of size 80 $\times$ 80 pixels, making the state-space very large. The state-space is, in fact, unnecessarily large as many input pixels aren't important. Again, this is not ATARI-specific but a general common problem. 
The \textit{focus-of-attention} (FoA) mechanism can reduce the amount of irrelevant pixels in the input and allow the agent to focus on certain parts of the screen while giving less attention to other parts \cite{foa}.\\

The addition of the FoA mechanism modifies the screens in a way that focuses the view of the agent on the relevant parts. For this, we introduce \textit{visual actions} that are used to move the center of attention at every time-step.
These actions enables the agent to be trained to focus its attention at interesting parts of the screen while simultaneously learning to play the Atari game. By introducing FoA we make sure that input pixels with no or little importance will, with enough training, be disregarded by the model. This will effectively reduce the size of the state-space.\\

The introduction of FoA makes the true game state only partially observable, which to some degree violates the Markov property. However, by moving the center of attention through the use of the visual actions the agent can access the relevant part of game state. This dynamic is meant to be similar to the human visual system where the human eyes are moving around and locating interesting parts in order to build up a mental map of a scene.

\subsection{Dual-head Architecture}
The model with the added FoA mechanism was created by duplicating the head in the A3C-LSTM model described earlier, see Fig. \ref{fig:a3c_foa}. One of the heads, the \textit{natural head}, controls the player movements or robotic behaviour within the game through \textit{natural actions} while the other head, the \textit{vision head}, controls the movements of the center of attention through \textit{visual actions}. The duplication of the head results in a model with four outputs: two softmax outputs for policies $\pi_{\theta^{nat}}^{nat}$, $\pi_{\theta^{vis}}^{vis}$ and two linear outputs for value functions $V_{\nu^{nat}}^{nat}$, $V_{\nu^{vis}}^{vis}$. 
Note that the four sets of parameters $\theta^{nat}$, $\theta^{vis}$, $\nu^{nat}$ and $\nu^{vis}$ share the parameters for the convolutional torso and the LSTM. This means that most parameters are shared and because of this the expressions are simplified as $\theta = \theta^{nat} = \theta^{vis} = \nu^{nat} = \nu^{vis}$ in this paper.\\

\begin{figure*}[t]
\includegraphics[width=0.8\textwidth]{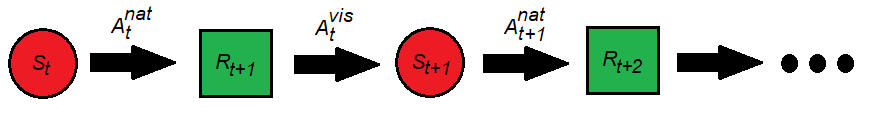}
\centering
\caption{Transition of A3C-LSTM with FoA.}
\label{fig:transition}
\end{figure*}

A transition from state $S_t$ to $S_{t+1}$ for this model is made through a natural action $A_t^{nat}$ and a simultaneous visual action $A_t^{vis}$. The reward $R_{t+1}$ from this transition depends only on the natural action $A_t^{nat}$, however. The visual action $A_t^{vis}$ doesn't directly affect the state of the game, but instead affects what part of the true game state that is accessible to the agent. This means that while $R_{t+1}$ is not affected by $A_t^{vis}$, the state $S_{t+1}$ is affected and thus also the following natural action $A_{t+1}^{nat}$ and the next reward $R_{t+2}$. To account for this behavior, the loss functions have to be modified. The loss functions for the natural head require no modifications but in the loss functions for the vision head the reward $R_{t+1}$ has to be replaced with the next reward $R_{t+2}$. Both $R_{t+1}$ and $R_{t+2}$ are thus needed and a transition can be represented by $(S_t, A_t^{nat}, A_t^{vis}, R_{t+1}, R_{t+2}, S_{t+1})$. A transition is illustrated in Fig. \ref{fig:transition}.

\begin{Figure}
\includegraphics[width=1\textwidth]{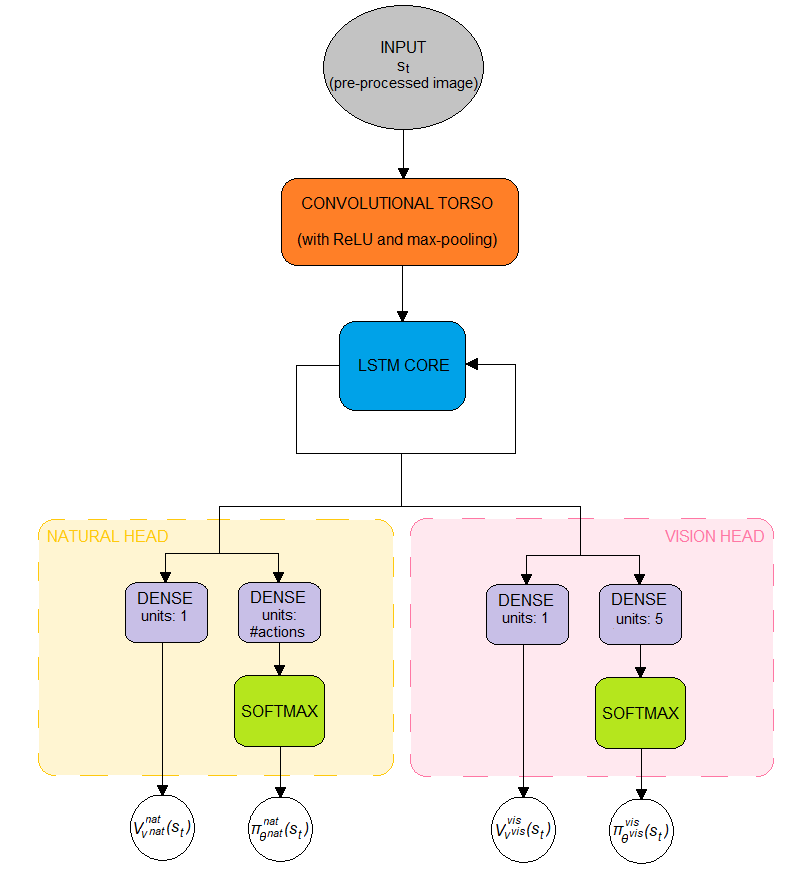}
\centering
\captionof{figure}{Overview of the A3C-LSTM with FoA.}
\label{fig:a3c_foa}
\end{Figure}

\subsection{Dual-head Loss Function}
With the dual-head architecture the total loss $L_{total}$ needs to be divided in four parts: $L_{value}^{nat}$, $L_{policy}^{nat}$, $L_{value}^{vis}$ and $L_{policy}^{vis}$. The value loss for the natural head can be defined as:
$$L_{value}^{nat} = \frac{1}{2} \sum\limits_{i=0}^{k-1} A_\theta^{nat}(s_{t+i})^2 \enspace,$$
$$A_\theta^{nat}(s_{t}) = (\sum_{i=0}^{k-1}\gamma^i R_{t+1+i} + \gamma^k V_\theta^{nat}(s_{t+k})) - V_\theta^{nat}(s_t)$$
where $A_\theta^{nat}(s_{t})$ is the non-generalized advantage estimation for the natural head. The next loss, the policy loss for the natural head, can be defined as:
$$L_{policy}^{nat} = \sum_{i=0}^{k-1}(-\mathrm{log}\pi_\theta^{nat}(a_{t+i}|s_{t+i})\hat{A}_{t+i}^{GAE,nat}-H^{nat}_{t+i}) \enspace,$$
$$H^{nat}_{t} = \beta H(\pi_\theta^{nat}(s_{t})) \enspace,$$
$$\hat{A}_{t}^{GAE,nat} = \sum\limits_{i=0}^{k-1}(\gamma\lambda)^i \delta^{nat}_{t+i} \enspace,$$
$$\delta^{nat}_{t} = R_{t+1} + \gamma V_\theta^{nat}(s_{t+1}) - V_\theta^{nat}(s_{t})$$
where $H^{nat}_{t}$ is the entropy term, $\hat{A}_{t}^{GAE,nat}$ is the generalized advantage estimation and $\delta^{nat}_{t}$ is the TD(0)-error. The losses for the vision head can be defined in similar manners, starting with the value loss:
$$L_{value}^{vis} = \frac{1}{2} \sum\limits_{i=0}^{k-2} A_\theta^{vis}(s_{t+i})^2 \enspace,$$
$$A_\theta^{vis}(s_{t}) = (\sum_{i=0}^{k-2}\gamma^i R_{t+2+i} + \gamma^k V_\theta^{vis}(s_{t+k})) - V_\theta^{vis}(s_t)\enspace .$$
Note how the reward has been replaced in the loss for the vision head. The last loss, the policy loss of the vision head, can be defined as:
$$L_{policy}^{vis} = \sum_{i=0}^{k-2}(-\mathrm{log}\pi_\theta^{vis}(a_{t+i}|s_{t+i})\hat{A}_{t+i}^{GAE,vis}-H^{vis}_{t+i}) \enspace,$$
$$H^{vis}_{t} = \beta H(\pi_\theta^{vis}(s_{t})) \enspace,$$
$$\hat{A}_{t}^{GAE,vis} = \sum\limits_{i=0}^{k-2}(\gamma\lambda)^i \delta^{vis}_{t+i} \enspace,$$
$$\delta^{vis}_{t} = R_{t+2} + \gamma V_\theta^{vis}(s_{t+1}) - V_\theta^{vis}(s_{t})\enspace.$$ 
The definitions above are slightly simplified as only one set of hyperparameters $\gamma, \beta, \lambda$ is considered instead of using one set $\gamma^{nat}, \beta^{nat}, \lambda^{nat}$ for the natural head and one set $\gamma^{vis}, \beta^{vis}, \lambda^{vis}$ for the vision head. Throughout this project $\gamma=\gamma^{nat}=\gamma^{vis}$, $\beta=\beta^{nat}=\beta^{vis}$ and $\lambda=\lambda^{nat}=\lambda^{vis}$ are used.\\

Pseudo-code for the A3C-LSTM with FoA can be found in Algorithm \ref{alg:foa}.

\begin{algorithm*}
\caption{Pseudo-code for the A3C-LSTM algorithm with focus-of-attention}\label{alg:foa}
\begin{algorithmic}
\State Initialize global network with parameters $\Theta$
\State Initialize global shared counter $T$
\State Choose amount of action-learners
\ForAll{action-learner threads}
\While{$T < T_{max}$}
\State Initialize local step-counter $t=0$
\State Synchronize local parameters $\theta=\Theta$
\State Get state $s_t$
\While{$t < t_{max}$ \textbf{and} $s_t$ is not terminal state}
\State Perform natural action $a_t^{nat}$ according to probabilistic policy $\pi_\theta^{nat}(s_t)$
\State Perform visual action $a_t^{vis}$ according to probabilistic policy $\pi_\theta^{vis}(s_t)$
\State Receive and save reward $R_{t+1}$ and new state $s_{t+1}$
\State $t \leftarrow t+1$
\State $T \leftarrow T+1$
\EndWhile
\State Compute total discounted rewards $R_{tot}^{nat}$ and $R_{tot}^{vis}$ through bootstrapping
\State Form estimate of advantage function for natural head $A_\theta^{nat}(s_t) = R_{tot}^{nat}-V_\theta^{nat}(s_t)$
\State Form estimate of advantage function for vision head using $A_\theta^{vis}(s_t) = R_{tot}^{vis}-V_\theta^{vis}(s_t)$
\State Compute generalized advantage estimations $\hat{A}_t^{GAE,nat}(s_t)$ and $\hat{A}_t^{GAE,vis}(s_t)$
\State Form losses $L_{value}^{nat}$, $L_{policy}^{nat}$, $L_{value}^{vis}$ and $L_{policy}^{vis}$
\State Form total loss $L_{total}$
\State Use loss to compute local gradient with respect to $\theta$
\State Update global $\Theta$ asynchronously using local gradient
\EndWhile
\EndFor
\end{algorithmic}
\end{algorithm*}

\subsection{Region of Interest Layer}
In order to use the game screens as input to the model without the focus-of-attention algorithm some pre-processing is needed. This pre-processing includes standard operations like cropping, resizing, gray-scale conversion and normalization. When the model with the FoA algorithm is used for an Atari environment an extra pre-processing layer is needed. This layer modifies the game screens in a way that focuses the attention of the agent at a specific region. This extra layer will be referred to as a \textit{Region of Interest} (RoI) layer in this paper. The different RoI layer investigated in the paper are presented below.\\

The simplest RoI layer defines the \textit{focal area}, the area which holds all the pixels that are visible to the agent, using a rectangle. To define said rectangle a focal point $(f_x, f_y)$, a focal width and a focal height are needed. The focal point marks the center of attention and can be moved through the visual actions of the agent. The resolution of the focal area also needs to be decided and is defined using a \textit{subsampling factor}. Due to the fact that all visible pixels are displayed in the same resolution, this RoI layer will be referred to as a \textit{Constant} resolution RoI layer.\\

Through a small change the focal area can be made more complex. By introducing more than one resolution within the focal area the RoI layer will better mimic the human visual system. This new layer still needs a focal point $(f_x, f_y)$ and will need three sets of focal widths, focal heights and subsampling factors. The first set will form a small rectangular area with high pixel resolution, the second set will form a medium-sized rectangular area with medium pixel resolution and the last set will form a large rectangular area with low pixel resolution. This RoI layer will be referred to as a \textit{Decreasing} resolution RoI layer. A comparison between the \textit{Constant} and the \textit{Decreasing} RoI layers can be found in Fig. \ref{fig:explain}.

\begin{Figure}
\includegraphics[width=0.48\textwidth]{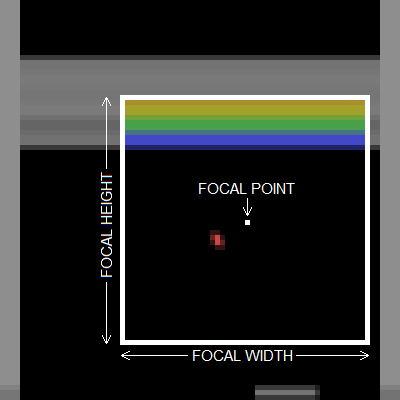}
\includegraphics[width=0.48\textwidth]{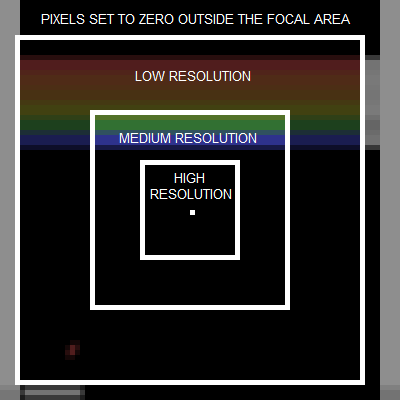}
\centering
\captionof{figure}{Comparison of \textit{Constant} and \textit{Decreasing} RoI layers.}
\label{fig:explain}
\end{Figure}

To further mimic the human visual system it is of interest to introduce peripheral vision to the agent. The peripheral vision of a human is wide but blurry, making it useful for noticing movements but ill-suited for making out details. Both the \textit{Constant} and the \textit{Decreasing} RoI layers are extended with something that simulates peripheral vision by making changes to the pixels outside the focal area. These pixels were previously set to zero, but are now replaced with a very low resolution representation of the game state. With this low resolution representation it is not possible to make out any detail, but it should be possible for the agent to detect changes. The resulting RoI layers will be referred to as \textit{Constant(P)} and \textit{Decreasing(P)}, respectively. Input states with corresponding visualizations for the \textit{Constant} RoI layer can be seen in Fig. \ref{fig:c_state} while input states and visualizations for the \textit{Decreasing(P)} RoI layer can be seen in Fig. \ref{fig:dp_state}.

\begin{Figure}
\includegraphics[width=0.85\textwidth]{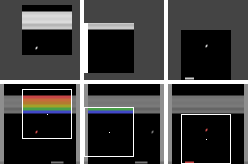}
\centering
\captionof{figure}{Input states (top) and corresponding visualizations (bottom) for the \textit{Constant} RoI layer.}
\label{fig:c_state}
\end{Figure}
\begin{Figure}
\includegraphics[width=0.85\textwidth]{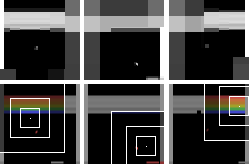}
\centering
\captionof{figure}{Input states (top) and corresponding visualizations (bottom) for the \textit{Decreasing(P)} RoI layer.}
\label{fig:dp_state}
\end{Figure}
\section{Experiments}
The experiments were carried out on different Atari 2600 games from the Arcade Learning Environment (ALE) suite \cite{ale}. The Atari games were implemented using OpenAI Gym \cite{openai} and while experiments were run on Pong, Breakout and Beam Rider. Due to limited space we will discuss only the Breakout results in this paper in detail but the other games performed similarly\footnote{The complete results for all three games can be found here: https://lup.lub.lu.se/student-papers/search/publication/9095697}.
A total of seven models were trained. The first one was a non-FoA model which would serve as a baseline.  The other six models use different RoI layers. A summary of the models can be seen in Table \ref{tab:sum} and the hyperparameters used can be seen in Appendix \ref{app:hyper}.

\begin{table*}
\caption{Model overview that includes focal area(s) in pixels, the subsampling factor(s), if a peripheral is used and the amount of total pixel-values that the agent has access to.}
\label{tab:sum}
\centering
\begin{tabular}{|c|c|c|c|c|} 
 \hline
 Model name & Focal area(s) & Subsampling factor(s) & Peripheral & Pixels-values \\
 \hline
 Non-FoA, sub1 & - & 1 & No & 6400\\ 
 Constant 50x50, sub1 & $50 \times 50$ & 1 & No & 2500\\ 
 Constant 50x50, sub2 & $50 \times 50$ & 2 & No & 625\\ 
 Constant 70x70, sub2 & $70 \times 70$ & 2 & No & 1225\\ 
 Decreasing 30-50-70 & $30 \times 30$, $50 \times 50$, $70 \times 70$ & 1, 2, 4 & No & 1450\\
 Constant(P) 70x70, sub2 & $70 \times 70$ & 2 & Yes & 1241\\ 
 Decreasing(P) 30-50-70 & $30 \times 30$, $50 \times 50$, $70 \times 70$ & 1, 2, 4 & Yes & 1466\\
 \hline
\end{tabular}
\end{table*}

\subsection{Constant Resolution Models}
First, the models using a \textit{Constant} RoI layer are compared with each other and with the non-FoA model. The performances of these different models can be seen in Fig. \ref{fig:constant}. From the figure it can be seen that the addition of the focus-of-attention algorithm generally deteriorates performance, particularly during the initial episodes where the agents don't seem to learn anything at all. This slow start is probably due to the fact that the size of the state-space is actually increased for untrained models when focus-of-attention is introduced. When applying the RoI layer the amount of actual game states remain unchanged and for each of these states there now exists many different input states based on the location of the focal point. It is not until the agent learns to control its focus-of-attention that the size of the state-space is effectively reduced. The reinforcement learning task also becomes more complex as the action-space grows. The effect of this can to some extent be seen when looking at the performance of the \textit{Constant 70x70, sub2} model in Fig. \ref{fig:constant}. This model barely learns anything during the initial episodes but after approximately 40000 episodes the performance improves at almost the same pace as the \textit{Non-FoA, sub1} model. \\

\begin{Figure}
\includegraphics[width=0.98\textwidth]{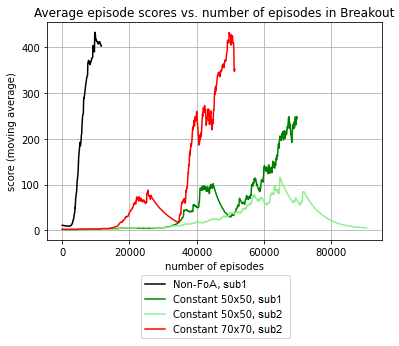}
\centering
\captionof{figure}{Performance in Breakout for \textit{Constant} resolution models.}
\label{fig:constant}
\end{Figure}

Fig. \ref{fig:constant} also shows that lowering the resolution for a $50 \times 50$ focal area deteriorates performance. Using a small focal area along with a lower resolution seems to remove too much information about the game state from the agent. Because of this the \textit{Constant 70x70, sub2} model is introduced. This model uses a larger focal area than the \textit{Constant 50x50, sub1} model but with a larger subsampling factor. Making these changes the model performance is improved, both when it comes to training speed and maximum score. Thus, it seems like the total size of the focal area is more important than the resolution for model performance. The amount of pixels visible to the agents can also be compared. The \textit{Constant 50x50, sub1} model uses a focal area of 2500 pixels with full resolution while the \textit{Constant 70x70, sub2} model uses a focal area of 4900 pixels where only a fourth of all pixels is kept due to the lowered resolution. This means that an agent of the \textit{Constant 50x50, sub1} model has access to 2500 pixels at any time while an agent of the \textit{Constant 70x70, sub2} model only has access to 1225 pixels. In other words, the \textit{Constant 70x70, sub2} model outperforms the \textit{Constant 50x50, sub1} model even though only approximately half the amount of pixels are visible to the agent.

\subsection{Decreasing Resolution Models}
Next, the \textit{Decreasing} RoI layer is introduced with the \textit{Decreasing 30-50-70} model and compared with the best performing model using the \textit{Constant} RoI layer, as can be seen in Fig. \ref{fig:decreasing}. The \textit{Decreasing 30-50-70} model uses an inner focal area of size $30 \times 30$ with a subsampling factor of 1, a middle focal area of size $50 \times 50$ with a subsampling factor of 2 and an outer focal area of size $70 \times 70$ with a higher subsampling factor of 4. This means that the two models are using the same total focal area. When computing how many pixels are visible to the agent, it can be shown that an agent of the \textit{Decreasing 30-50-70} model has access to 1450 pixels while an agent of the \textit{Constant 70x70, sub2} model has access to 1225 pixels. The amount of pixels visible to agents of the two models are thus almost the same. \\

The performances of the two models are quite similar even though increasing the resolution in a small focal area should improve performance, as seen when comparing the models using the \textit{Constant} RoI layer in Fig. \ref{fig:constant}. However, as no boost in performance can be seen by increasing the resolution of the innermost focal area, this effect seems to be negated by the lowered resolution in the outermost focal area. It should nevertheless be noted that the low resolution outer focal area does add valuable information to the model. This can be seen by comparing the performance of the \textit{Decreasing 30-50-70} model in Fig. \ref{fig:decreasing} with the performance of the \textit{Constant 50x50, sub1} model in Fig. \ref{fig:constant}. The \textit{Decreasing 30-50-70} can be seen to perform better than the \textit{Constant 50x50, sub1} model and the only thing added to the RoI layer for this model is an extra low resolution padding (the outer focal area).\\

The different resolutions of the \textit{Decreasing 30-50-70} model can possibly make the training process more challenging for the agent. When only using one subsampling factor a certain game object always look the same but when using three different subsampling factors the same game object can take on multiple shapes. A ball can look like a sharp dot if located in the center of the focal area but it can also look like a blurry blob if located further from the focal point. Also, at some point in time a certain area of the screen can be given in high resolution and at another point in time this same area can be given in a lower resolution. In other words, the multiple resolutions of the \textit{Decreasing} RoI layer introduces a new challenge for the agent.

\begin{Figure}
\includegraphics[width=0.98\textwidth]{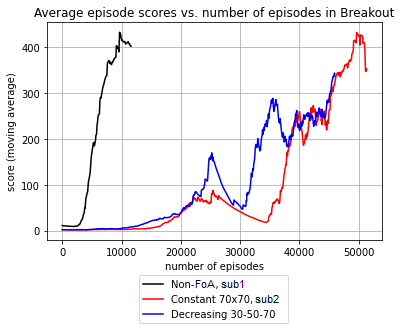}
\centering
\captionof{figure}{Performance in Breakout for \textit{Decreasing} resolution models.}
\label{fig:decreasing}
\end{Figure}

\subsection{Peripheral Vision}
In the final models, peripheral vision is introduced. The models with peripheral vision, \textit{Constant(P) 70x70, sub2} and \textit{Decreasing(P) 30-50-70} are compared with their non-peripheral counterparts \textit{Constant 70x70, sub2} and \textit{Decreasing 30-50-70} in Fig. \ref{fig:peripheral}. In these new models the zero-valued background is replaced with a $5 \times 5$ resolution version of the game state. The peripheral vision makes a maximum of 16 extra pixels visible to the agent as a few of the 25 background pixels always will be hidden by the focal area. The few extra pixels do, however, improve performance significantly as both the \textit{Constant(P) 70x70, sub2} model and the \textit{Decreasing(P) 30-50-70} model are faster to train than their counterparts. This improvement can be explained with the help of the heat maps in Fig, \ref{fig:heatmap}. In the heat maps it can be seen that the attention of the agent usually is focused on the lower half of the screen. This means that without the peripheral vision the agent will not know where the remaining brick are located. However, when adding the low resolution background the agent can make out in which area of the screen the remaining bricks are located and thus, more easily, progress in the game. The peripheral thus adds valuable information to the model which is in accordance with the previous observation that the total size of the focal area is more important than the resolution.

\begin{Figure}
\includegraphics[width=0.98\textwidth]{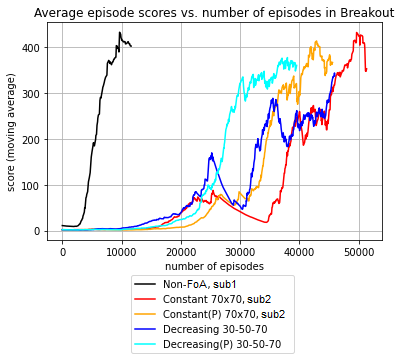}
\centering
\captionof{figure}{Performance in Breakout for models with peripheral vision.}
\label{fig:peripheral}
\end{Figure}

\subsection{Agent Behavior}
The behavior of the agent can be seen in the heat maps in Fig. \ref{fig:heatmap}. The left heat map shows how a trained agent of the \textit{Constant 70x70, sub2} model uses its visual actions to move its focal point while the right heat map shows the corresponding behavior for the \textit{Decreasing(P) 30-50-70} model. From the heat maps it can be seen that the agent does not move its focal point very much. In fact, the focal point is mostly kept centered at the bottom half of the screen, which is the most important area of the screen when playing Breakout. The lack of focal point movement is likely due to the large size of the focal areas; even without movement the majority of the screen will be visible to the agent. It should however be noted that there are more focal point movement for the better performing \textit{Decreasing(P) 30-50-70} model. This indicates that even though a trained model will keep its focus-of-attention in a specific area when playing the movement of the focal point is still of importance.

\begin{Figure}
\includegraphics[width=0.48\textwidth]{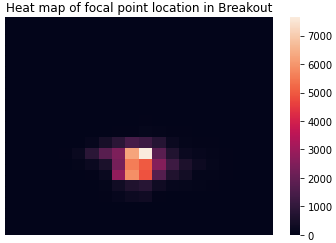}
\includegraphics[width=0.48\textwidth]{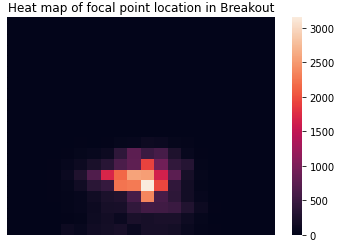}
\centering
\captionof{figure}{Heat map for focal point movement for the \textit{Constant 70x70, sub2} and \textit{Decreasing(P) 30-50-70} models.}
\label{fig:heatmap}
\end{Figure}
\section{Conclusions}
In this paper the A3C-LSTM network was extended with a focus-of-attention mechanism and trained on the Atari 2600 game Breakout. An agent was required to both learn to control its movements and its focus-of-attention. Region of Interest layers with constant resolution, decreasing resolution and with peripheral vision were used. In general, the introduction of the focus-of-attention task slowed down the training process. The drop in performance is probably due to the larger action-spaces and the larger state-spaces of the untrained FoA models. \\
First, a model using a \textit{Constant} resolution RoI layer with a focal area of size $50\times 50$ was shown to achieve high scores on Breakout. By then introducing a RoI layer with focal area of size $70 \times 70$ with lower resolution is could be shown that in these games the size of the focal area is more important than the resolution of the input state. High resolution pixels do not seem to hold much more valuable information than lower resolution pixels but the lower resolution pixels hold more valuable information than no pixels at all. \\
When changing the $70 \times 70$ \textit{Constant} RoI layer into a \textit{Decreasing} resolution RoI layer, still with a focal area of size 70 × 70 but with varying resolution, the performance was not improved. The higher resolution closest to the focal point does not make up for the loss of information related to the lower resolution far from the focal point. The \textit{Decreasing} resolution RoI layer does, however, perform better than the $50 \times 50$ \textit{Constant} RoI layer. The addition of the extra low resolution area does, in other words, improve the performance significantly. Again, it would seem that in these games the size of the focal area is more important than the use of a high resolution. \\
The addition of the peripheral vision only gives an agent access to a few extra pixel-values but affects the performance quite a lot. When the peripheral is added to the two previously mentioned RoI layers the performance is improved. In both these cases the peripheral seems to provide important information to the agent. \\
In summary, we were able to demonstrate working gaze control based on log-polar-like images. Pong and Breakout did not seem to take much advantage from the gaze control. On the other hand, a low-resolution broad field of view proved important. The heat maps in Fig. \ref{fig:heatmap} reflects the most important areas of the screen when playing Breakout, and one can see that the gaze did not move much. Still, without the gaze control and only a low-resolution image the performance was much worse.
We are now exploring our approach on object detection, and we expect that this task would profit more from of the high-resolution center area.  


\section*{Acknowledgements}
This was was supported by the Wallenberg Autonomous Systems Program (WASP), Sweden. 
The computations were enabled by the supercomputing resource Berzelius provided by National Supercomputer Centre at Linköping University and the Knut and Alice Wallenberg foundation.
\printbibliography
\appendix
\section{Appendix}

\subsection{Hyperparameters}\label{app:hyper}

\begin{table}[H]
\caption{Hyperparameters used for the models in the paper.}
\label{tab:hyper}
\centering
\begin{tabular}{|c|c|} 
 \hline
 \textbf{Hyperparameter} & \textbf{Value} \\
 \hline
 \multicolumn{2}{|c|}{Atari pre-processing}\\
 \hline
 Frame size & $80 \times 80$ \\
 Gray-scale & Yes \\
 Cropped images & Yes \\
 Normalized images & Yes \\
Frame skip & 4 \\
Max-pool screens & Yes \\
Frame stack & No \\
Initial no-operations & $[0, 30]$ \\
Clip rewards to \{$-1, 0, +1$\} & Yes \\
Loss of life is terminal state & Yes \\
Fire on reset & Yes \\
Number of actions & Game specific \\
 \hline
 \multicolumn{2}{|c|}{Training}\\
 \hline
 Learning rate, $\alpha$ & $1.0 \cdot 10^{-4}$ \\
 Optimizer & Shared Adam (AMSGrad) \\
Action-learner threads & 32 \\
Max number of steps, $t_{max}$ & 20 \\
Discount factor, $\gamma$ & 0.99 \\
GAE parameter, $\lambda$ & 0.92 \\
Entropy term, $\beta$ & 0.01 \\
 \hline
\end{tabular}
\end{table}

\end{document}